\documentclass[letterpaper, 10 pt, conference]{ieeeconf}  

\IEEEoverridecommandlockouts                              

\usepackage{cite}
\usepackage{amsmath,amssymb,amsfonts}
\usepackage{algorithmic}
\usepackage{graphicx}
\usepackage{textcomp}
\usepackage[dvipsnames]{xcolor}
\usepackage{multirow}
\usepackage{makecell}
\usepackage{tikz}
\usepackage{url}
\usepackage{hyperref}
\usepackage{subcaption}
\usepackage{color, colortbl}
\usepackage{xcolor}
\definecolor{LightBlue}{RGB}{212, 250, 252} 

\usepackage{tikz}
\newcommand\copyrighttext{%
  \footnotesize \textcopyright 2025 IEEE.  Personal use of this material is permitted.  Permission from IEEE must be obtained for all other uses, in any current or future media, including reprinting/republishing this material for advertising or promotional purposes, creating new collective works, for resale or redistribution to servers or lists, or reuse of any copyrighted component of this work in other works.}
\newcommand\copyrightnotice{%
\begin{tikzpicture}[remember picture,overlay]
\node[anchor=south,yshift=10pt] at (current page.south) {\fbox{\parbox{\dimexpr\textwidth-\fboxsep-\fboxrule\relax}{\copyrighttext}}};
\end{tikzpicture}%
}

\title{\LARGE \bf
Fool the Stoplight: Realistic Adversarial Patch Attacks on Traffic Light Detectors
}

\author{Svetlana Pavlitska$^{1,2}$, Jamie Robb$^{2}$, Nikolai Polley$^{1}$, Melih Yazgan$^{2}$, and J. Marius Zöllner$^{1,2}$
\thanks{$^{1}$ Karlsruhe Institute of Technology (KIT), Kaisterstr. 12 Karlsruhe Germany. {\tt\small \{prename.surname\}@kit.edu}.}
\thanks{$^{2}$ Department of Technical Cognitive Systems, FZI Research Center for Information Technology, Germany.
	{\tt\small \{surname\}@fzi.de}.}%
}

\begin{document}

\maketitle
\copyrightnotice
\thispagestyle{empty}
\pagestyle{empty}

\begin{abstract}

Realistic adversarial attacks on various camera-based perception tasks of autonomous vehicles have been successfully demonstrated so far. However, only a few works considered attacks on traffic light detectors. This work shows how CNNs for traffic light detection can be attacked with printed patches. We propose a threat model, where each instance of a traffic light is attacked with a patch placed under it, and describe a training strategy. We demonstrate successful adversarial patch attacks in universal settings. Our experiments show realistic targeted red-to-green label-flipping attacks and attacks on pictogram classification. Finally, we perform a real-world evaluation with printed patches and demonstrate attacks in the lab settings with a mobile traffic light for construction sites and in a test area with stationary traffic lights. Our code is available at \url{https://github.com/KASTEL-MobilityLab/attacks-on-traffic-light-detection}.

\end{abstract}

\section{Introduction}

Traffic light detection in automated vehicles typically combines multiple modalities to ensure accuracy and robustness. Deep learning-based computer vision is a crucial component, usually enhanced with map and GPS data~\cite{fairfield2011traffic,possatti2019traffic}, vehicle-to-infrastructure communication, and sensor fusion techniques. Camera-based detection relies primarily on convolutional neural networks (CNNs), whereas detections can be improved with temporal smoothing to ensure detection stability and avoid flickering errors. 

CNNs are, however, known to possess inherent brittleness. A deliberately generated input can evoke a misprediction, the so-called \textit{adversarial attack}~\cite{szegedy2013intriguing, goodfellow2014explaining}. Adversarial attacks can even cause malfunctioning of deep neural networks in the real world. For this, a printed locally-bound visible adversarial noise (an \textit{adversarial patch}) can be physically added to a scene perceived with a camera~\cite{brown2017adversarial}. Successful attacks on various camera-based perception tasks have been demonstrated, including object detection, semantic segmentation~\cite{nesti2022evaluating}, traffic sign recognition~\cite{eykholt2018robust,wei2023adversarial}, and steering angle prediction\cite{pavlitskaya2020feasibility}. However, to the best of our knowledge, no patch-based attacks on traffic light detectors have been demonstrated so far. Described attacks on traffic light recognition models are scarce; the attacks either rely on invisible noise instead of a patch~\cite{wan2020effects,yang2024sitar} or are performed with a laser~\cite{yan2022rolling,bhupathiraju2024vulnerability}. 

\begin{figure}[t]
\centering
\begin{subfigure}[t]{0.45\linewidth}
    \includegraphics[width=\textwidth]{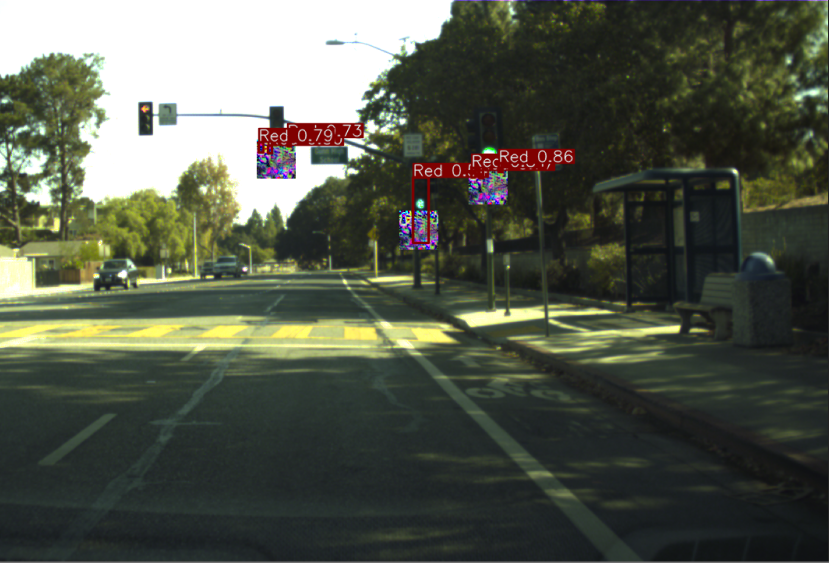}
    \caption{\texttt{BSTLD} dataset}
\end{subfigure}
\begin{subfigure}[t]{0.41\linewidth}
    \includegraphics[width=\textwidth]{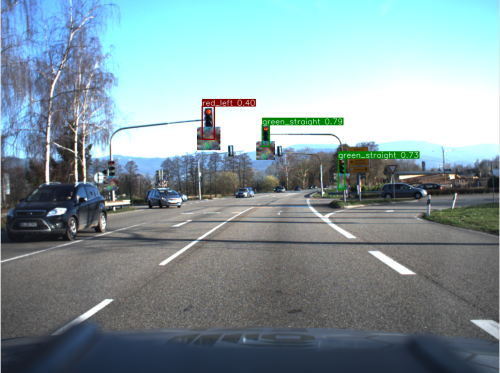}
    \caption{\texttt{HDTLR} dataset}
\end{subfigure}

\begin{subfigure}[t]{0.45\linewidth}
    \includegraphics[width=\textwidth]{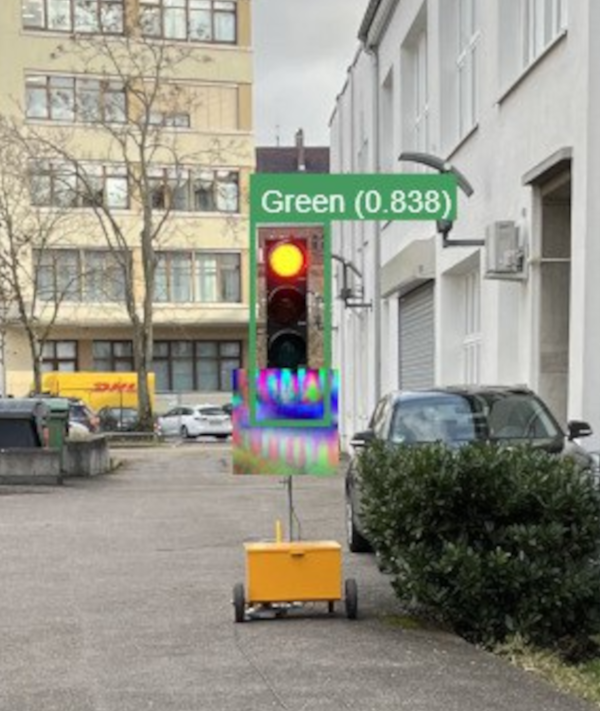}
    \caption{Mobile traffic light}
\end{subfigure}
\begin{subfigure}[t]{0.45\linewidth}
    \includegraphics[width=\textwidth]{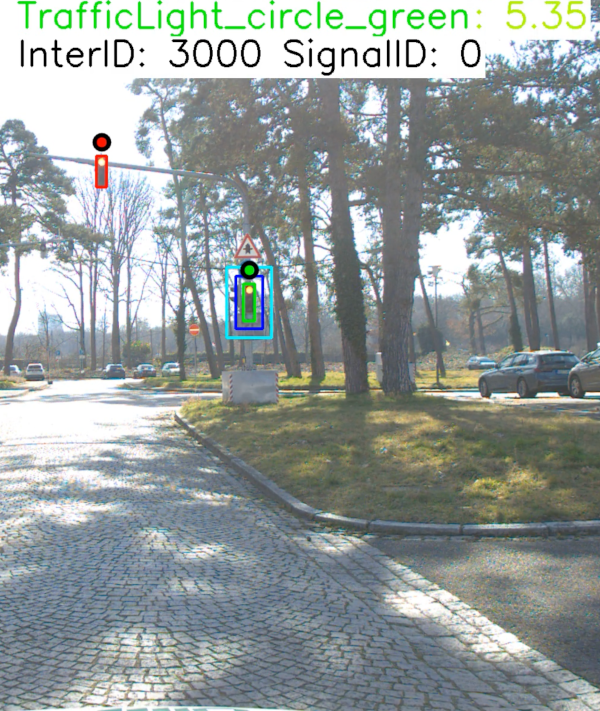}
    \caption{Stationary traffic light}
\end{subfigure}
    \caption{Examples of proposed attacks with an adversarial patch placed under the traffic light, evaluated in the digital (a-b) and real-world (c-d) settings.}
    \label{fig:tlr}
\end{figure}

Our work aims to close this gap and proposes patch-based attacks on traffic light detection models. Traffic lights present several challenges for a patch-based attack. First, the appearance of an object under attack changes depending on the traffic light state. Next, traffic lights are not as easily accessible as road signs, making sticker-based attacks and replacing the whole object with a printed pattern unfeasible. We propose to put a patch directly under the traffic light, thus mimicking posters usually attached to traffic light poles (cf. \cite{zhou2020deepbillboard,pavlitskaya2020feasibility}). We further propose a loss function to enforce the correct localization of a bounding box (BBox) of a traffic light with a simultaneous label flip. The described attacks are evaluated on four state-of-the-art datasets and in real-world settings (see Figure~\ref{fig:tlr}). For the latter, we use a mobile traffic light system for construction sites and a more complex setting with a stationary traffic light in the Test Area Autonomous Driving Baden-Württemberg.

Our previous work~\cite{polley2023tld} has extensively evaluated different generic object detectors on four state-of-the-art traffic light datasets and in road tests with an automated vehicle. We adopt the same setup for the evaluation of attacks in this work. In particular, we use YOLOv7 and YOLOv8 for the evaluation on four datasets. For the real-world deployment, YOLOv8x trained on the largest dataset is used. We analyze vulnerability to attacks across models and datasets.

\newpage
\section{Related Work}

\subsection{Traffic Light Detection}
Traffic light detection from camera images can be performed using traditional computer vision approaches (color thresholding, edge detection, shape-based detection, template matching) or machine learning approaches. For the latter, handcrafted features can be used in combination with traditional classifiers like SVMs or decision trees~\cite{franke2001door, chung2002vision, lindner2004robust, nienhuser2010visual}. Using deep neural networks allows for a more robust traffic light detection, whereas CNNs, semantic segmentation networks, or Region Proposal Networks can be applied.

In our recent work~\cite{pavlitska2023traffic} three categories of CNN-based approaches were analyzed: (1) modifications of generic detectors like YOLO~\cite{redmon2016you} or SSD~\cite{liu2016ssd} for the traffic light detection~\cite{jensen2016vision, muller2018detecting, pon2018hierarchical, bach2018deep, aneesh2019real, gokul2020comparative, yan2021end, liu2023traffic}, (2) multi-stage approaches encompassing rule- and CNN-based components~\cite{han2019real, wang2022traffic, naimi2021fast}, and (3) task-specific single-stage methods~\cite{weber2016deeptlr, weber2018hdtlr, john2014traffic, john2015saliency}. The analysis has shown that methods belonging to the first group are predominant, with YOLO being the most used architecture. Furthermore, the shortage of available open-source implementations and the focus on classifying traffic light states (red, green, yellow) rather than arrows has been revealed.

\subsection{Adversarial Patch Attacks}

Since the inherent vulnerability of deep neural networks has been discovered~\cite{goodfellow2014explaining, szegedy2013intriguing}, a significant effort has been put into performing adversarial attacks in the real world. The standard adversarial attack approach consists of adding small-magnitude perturbations.
A further step towards realistic attacks is an \textit{adversarial patch}, where the adversarial perturbation is applied not to all input image pixels but within a specified image region~\cite{brown2017adversarial}. To perform an attack in the real world, an adversarial patch should be trained in a universal manner~\cite{moosavi2017universal}. Universal adversarial patch attacks have been successfully applied to various computer vision tasks, including object detection~\cite{karmon2018laVAN,pavlitskaya2022adversarial,pavlitskaya2022feasibility}, semantic segmentation~\cite{nesti2022evaluating}, and steering angle prediction~\cite{pavlitskaya2020feasibility}.

While our attack strategy is inspired by previous works describing patch-based attacks, traffic light detection yields several interesting challenges for attack generation. In particular, traffic lights are characterized by a relatively small size in the resulting image. On the other hand, unlike traffic signs, which are also relatively small and have been extensively studied, changing the whole traffic light area or putting adversarial stickers~\cite{ye2021patch,wei2023adversarial} on the salient areas of a traffic light is not feasible. Furthermore, for a successful attack, the same patch should be applied to the same instance of a traffic light showing different states. In contrast, existing work on adversarial patches rarely optimizes for changes in an object under attack.

\subsection{Attacks on Traffic Light Recognition}
Attempts to disrupt the performance of traffic light recognition have so far included attacks via network intrusion~\cite{ozarpa2021cyber,ghena2014green}, GPS spoofing~\cite{tang2021fooling} or traffic congestion~\cite{chen2018exposing,oza2020secure}. Another line of research focuses on optical attacks on traffic light recognition models. Wang et al.~\cite{wang2021icansee} used the infrared light LEDs in the traffic light to create a fake invisible red signal. Yan et al.~\cite{yan2022rolling} proposed an attack generating a laser beam and pointing it on a camera in an automated vehicle to cause a red light to be recognized as green and vice versa. For this, the rolling shutter effect in cameras is exploited under the assumption that most cameras in autonomous vehicles use rolling shutters. The experiments were performed using YOLOv4 and Apollo~\cite{apollo} on the BDD100K dataset~\cite{BDD100k} with two classes (red and green) in a simulation as well as in the real world, both stationary and in motion. The demonstrated attack successfully fools traffic light recognition, whereas the green-to-red attack is stronger. The proposed attack, however, requires synchronization with the camera in a vehicle and tracking of a car with a camera under attack.

Another laser-based attack by Bhupathiraju et al.~\cite{bhupathiraju2024vulnerability} uses a laser pointed at a traffic light. An attacker can perform light injection from a distance of up to 25m. The authors focus only on classifying the traffic light states, not detection. For evaluation, traffic light recognition in Apollo~\cite{apollo} (a CNN classifying four states), Autoware~\cite{kato2018autoware} (MobileNet~\cite{howard2017mobilenets} classifying 11 traffic light states) and an Inception-V3~\cite{szegedy2016rethinking} model trained on the LISA dataset~\cite{jensen2016vision} were chosen. The authors measured the success rate on the test data and during a proof-of-concept evaluation in an indoor scenario. For the latter, up to 100\% of attack success in low-speed vehicle scenarios was reached.

Significantly less attention has been paid to adversarial attacks on neural networks for traffic light recognition. One of the few works is by Wan et al.~\cite{wan2020effects}. Here, boundary~\cite{brendel2017decision}, spatial~\cite{engstrom2018rotation}, one-pixel~\cite{su2019one}, and C\&W attacks~\cite{carlini2017towards} on VGG models~\cite{simonyan2015very} for traffic light classification were performed. Models were trained on 477 synthetic images collected in CARLA~\cite{dosovitskiy2017carla} with three classes (red, yellow, and green). Attacks on single images using small invisible perturbations were evaluated. While this work was the first to approach this topic, the threat model is not feasible for the real world, as no detectors or real-world data was used, and the generated attack pattern is not universal but image-specific.

Another work in this direction is by Yang et al.~\cite{yang2024sitar}, who evaluated PGD~\cite{madry2017towards} attacks on the traffic light recognition pipeline in Apollo. The pipeline involves the identification of regions of interest (ROIs) based on the geographical information of traffic light locations. Then, each ROI is processed by Faster RCNN to output BBoxes. Finally, another CNN is used to predict the color of each BBox. The proposed attack perturbs pixels in an entire input image and is considered successful if a traffic light detection is suppressed, a BBox does not overlap with the ground truth, or the traffic light color is mispredicted. The proposed attack was evaluated on 400 selected Apollo images. 

In summary, neither of these two works regarding adversarial attacks on traffic light recognition performed physically realizable attacks with adversarial patches, nor performed evaluation in the real world. These attacks also use models trained to distinguish a restricted set of classes (red and green) and demonstrate the attack on a selected subset of images.

\newpage
\section{Method}
This section describes the proposed attack's threat model and patch generation algorithm for the proposed attack.

\subsection{Threat Model}

\begin{figure}[t]
\centering
\begin{subfigure}[t]{\linewidth}
\centering
    \includegraphics[width=0.8\textwidth]{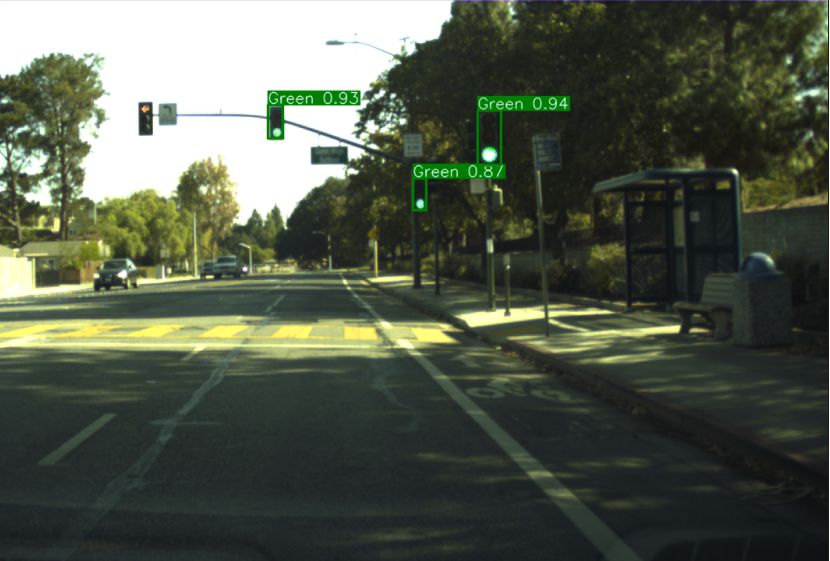}
  	\caption{No attack}
\end{subfigure}
\begin{subfigure}[t]{\linewidth}
\centering
    \includegraphics[width=0.8\textwidth]{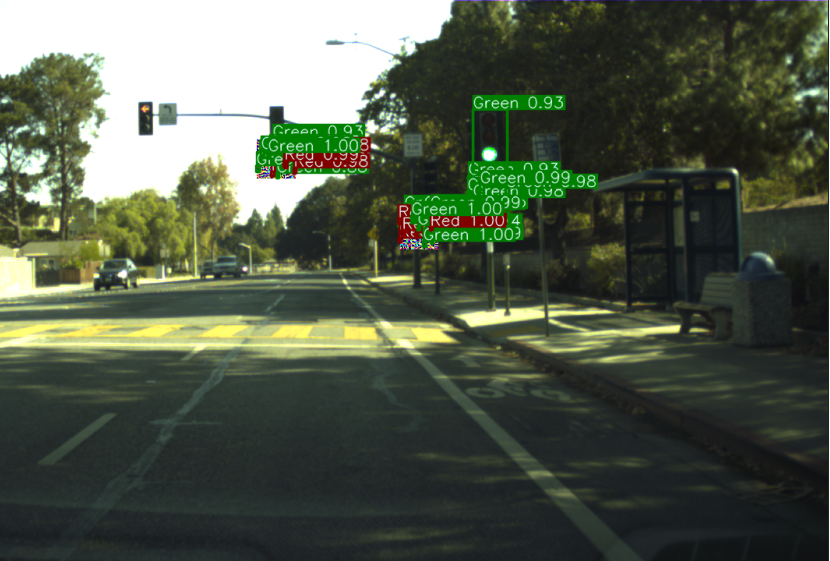}
  	\caption{Object fabrication attack}
\end{subfigure}
\begin{subfigure}[t]{\linewidth}
\centering
    \includegraphics[width=0.8\textwidth]{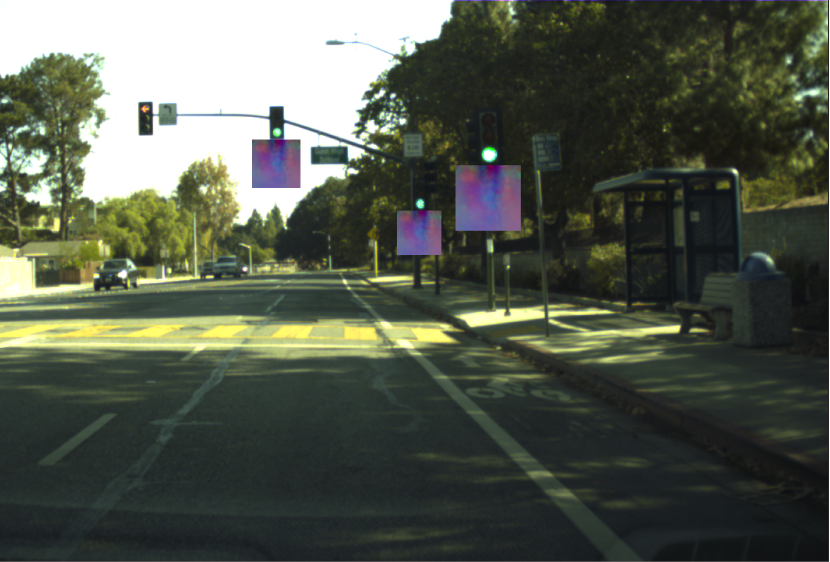}
  	\caption{Object vanishing attack}
\end{subfigure}
    \caption{Different attack types, exemplary for an image from \texttt{BSTLD}, predictions by YOLOv7x.}
    \label{fig:attack-types}
\end{figure}

We consider white-box attacks (i.e., an attacker has access to the network architecture, weights, and data) in the universal setting~\cite{moosavi2017universal} (i.e., a single noise pattern can attack all data). Three types of attacks on object detection models generally exist (cf.~\cite{chow2023tog}): (1) untargeted object vanishing attack enforcing false negatives, (2) untargeted object fabrication attack enforcing false positives,  and (3) targeted object mislabeling attack. Attacks of the first two groups on traffic light detection models are possible (see Figure~\ref{fig:attack-types}) but present a less realistic scenario: the object vanishing attack can be bypassed with a stored map of traffic light locations, whereas the object fabrication attack yields implausible predictions. 

To focus on realistic threat scenarios, we only consider targeted object mislabeling attacks aiming at two attacker's goals (cf.~\cite{yan2022rolling}): 1) \textit{red}$\to$\textit{green}: make a vehicle move while it should stop at a red light, and 2) \textit{green}$\to$\textit{red}: make a vehicle stop while it should move at a green light. Both scenarios can cause intersection accidents, while we focus on the latter as we rate it's attack criticality higher. 


\subsection{Patch Generation}
We formulate the attack goal for a given traffic light detector $D$, an input image $x$ with a set of ground truth predictions $Y=\{(y_{coord}, y_{conf\_score}, y_{class})\}$ (a set of BBox coordinates, a confidence score, and a label per detection) and a target class mapping function $m: Y \to Y$. The attack goal is to find an adversarial patch $p$, s.t. $D(A(p,x,l,t)) = Y'$, where $Y' = \{(y_{coord}, {y'_{conf\_score}}, m(y_{class})\}$ with the new target class $m(y_{class})$ and a new confidence score $y'_{conf\_score}$ obtained during inference. The patch should thus preserve the localization but swap the classification labels.

The patch application operator $A(p, x, l, t)$~\cite{brown2017adversarial} applies the transformations $t$ to the patch $p$ and then places the transformed patch $p$ to the image $x$ at location $l$. During training, an adversarial patch is placed directly under a ground-truth BBox of each traffic light if it fits in image boundaries. The transformation $t$ rescales a patch $p$ to be two to three times wider than a corresponding BBox. This scaling factor for a patch was determined experimentally. 

To generate a patch for a \textit{red}$\to$\textit{green} attack, we use the PGD-10 attack~\cite{madry2017towards} with the following loss function:

\begin{equation*}
\label{eq:base-loss}
\mathcal{L}_{\text{base}} = \alpha \mathcal{L}_{\text{cls}} + \beta \mathcal{L}_{\text{bbox}} +  \gamma \mathcal{L}_{\text{tv}} + \delta \mathcal{L}_{\text{green\_sup}}, 
\end{equation*}

where $\mathcal{L}_{\text{cls}} $ and $\mathcal{L}_{\text{bbox}}$ are YOLO classification and localization losses, respectively. $\mathcal{L}_{\text{tv}}$ is the total variation loss~\cite{mahmood2016accesorize} enforcing smoother transitions between neighboring pixels. $\mathcal{L}_{\text{green\_sup}}$ loss minimizes the number of green patch pixels and thus prevents unintended object fabrication. This loss calculates a penalty based on the amount of green pixels after applying a Gaussian blur to the green channel. Analogously, red pixels can be suppressed for a \textit{green}$\to$\textit{red} attack.

Patch training was performed using the Adam optimizer~\cite{kingma2014adam} with a learning rate set to 0.05. Since a patch should be placed under each instance of a traffic light, we perform a patch update for each relevant ground truth BBox in an image for 10 PGD steps.

\subsection{Patch Transformations for Real-world Attacks}
For real-world attacks, we additionally applied the expectation over transformations approach by Athalye et al.~\cite{athalye2018synthesizing}. During training, a transformation function is sampled from a predefined set. As in~\cite{lee2019physical}, this consists of a rotation around the x-axis and y-axis in the interval from -5$^{\circ}$ to 5$^{\circ}$, a rotation around the z-axis in the interval from -10$^{\circ}$ to 10$^{\circ}$, a scaling of the brightness with a factor in [0.4, 1.6], and a random translation within the area padded with 10 pixels.


\newpage
\section{Evaluation in the Digital Settings}
\label{ref:sec_digital}

We first evaluate the proposed attack on the test images from four state-of-the-art datasets.

\subsection{Experimental Setup} 
We used the best-performing models from our previous work~\cite{polley2023tld}\footnote{\url{https://github.com/KASTEL-MobilityLab/traffic-light-detection}}: YOLOv7~\cite{wang2023yolov7} and YOLOv8~\cite{jocher2023yolov8} trained on \texttt{BSTLD}, \texttt{LISA}, \texttt{HDTLR} and \texttt{DTLD}. Patches were trained on the corresponding train subsets and evaluated on test subsets.

The number and size of traffic lights per image varies heavily across datasets (see Table~\ref{tab:datasets}, cf.~\cite{fernandez2018deep}). Whereas 1-3 patch updates per image are performed for \texttt{BSTLD}, \texttt{LISA}, and \texttt{HDTLR}, on average, seven updates are required for \texttt{DTLD}. Furthermore, since BBoxes are relatively small in \texttt{BSTLD}, patches are heavily rescaled for each update. Depending on the dataset, the patch is initially set to $50\times50$ to $100\times100$ pixels.


\begin{table}[h]
\caption{Overview of the used datasets.}
\label{tab:datasets}
\begin{center}
\resizebox{1.0\linewidth}{!}{
    \begin{tabular}{|r|c|c|c|c|c| }
    \hline
    \textbf{Dataset}& \textbf{Classes} & \textbf{Train / Test} & \textbf{Avg} & \textbf{Max} & \textbf{Avg} \\ 
    & &  & \textbf{BBox} & \textbf{BBox} & \textbf{$\#$} \\
    &  &   & \textbf{size} & \textbf{size} & \textbf{BBoxes} \\
    \hline
     
    \texttt{BSTLD} & 4 & 5,093 / 8,334  & 11$\times$23 & 94$\times$154& 1.85\\\hline
    \texttt{LISA} & 7 & 12,775 / 3,348 & 35$\times$56 & 144$\times$194& 2.96 \\ \hline
    \texttt{HDTLR}  & 16 & 4,188 / 1,098  &  20$\times$55 & 95$\times$262& 2.15\\ \hline
    \texttt{DTLD}  & 20 & 28,525 / 12,453 & 16$\times$51 & 185$\times$552& 7.02\\ \hline
    \end{tabular}
}
\end{center}

\end{table}


\subsection{Analysis}

\begin{figure}[h]
\centering
\begin{subfigure}[t]{0.2\linewidth}
    \includegraphics[width=\textwidth]{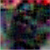}
\end{subfigure}
\begin{subfigure}[t]{0.2\linewidth}
    \includegraphics[width=\textwidth]{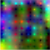}
\end{subfigure}
\begin{subfigure}[t]{0.2\linewidth}
    \includegraphics[width=\textwidth]{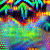}
\end{subfigure}
\begin{subfigure}[t]{0.2\linewidth}
    \includegraphics[width=\textwidth]{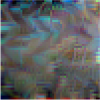}
\end{subfigure}
    \caption{Examples of generated patches for states and arrows.}
    \label{fig:patches}
\end{figure}

\begin{figure}[t]
\centering
\begin{subfigure}[t]{0.72\linewidth}
    \includegraphics[width=\textwidth]{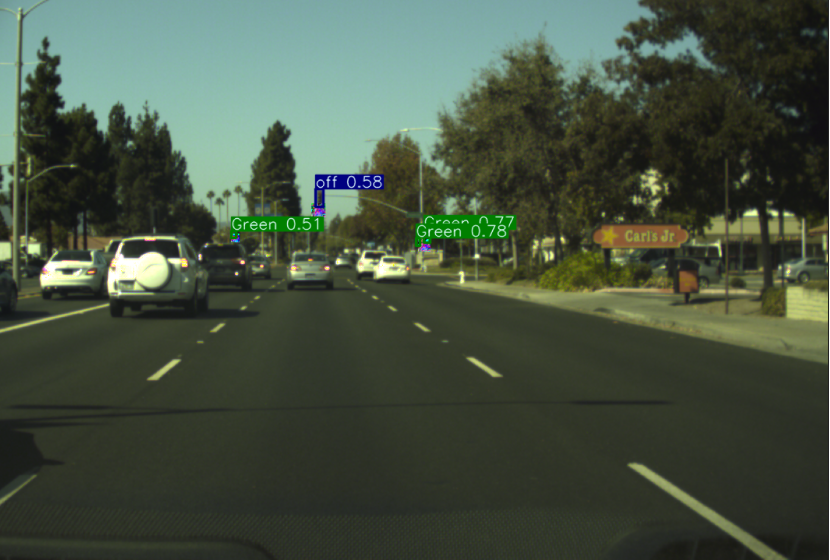}
  	\caption*{\texttt{BSTLD}}
\end{subfigure}

\begin{subfigure}[t]{0.72\linewidth}
    \includegraphics[width=\textwidth]{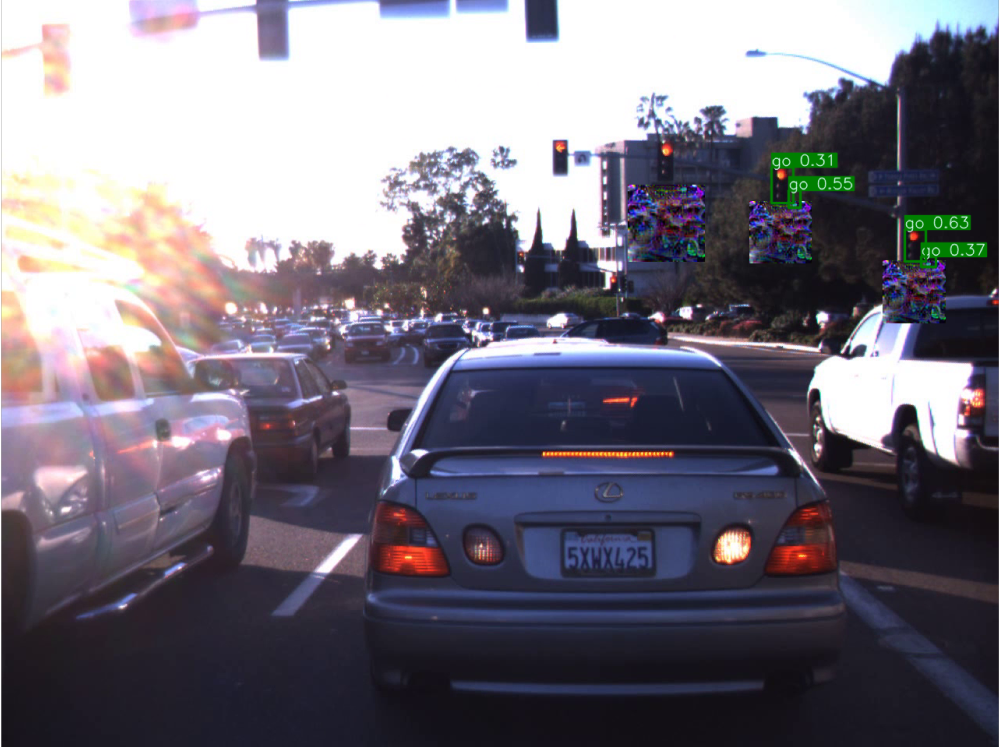}
  	\caption*{\texttt{LISA}}
\end{subfigure}

\begin{subfigure}[t]{0.72\linewidth}
    \includegraphics[width=\textwidth]{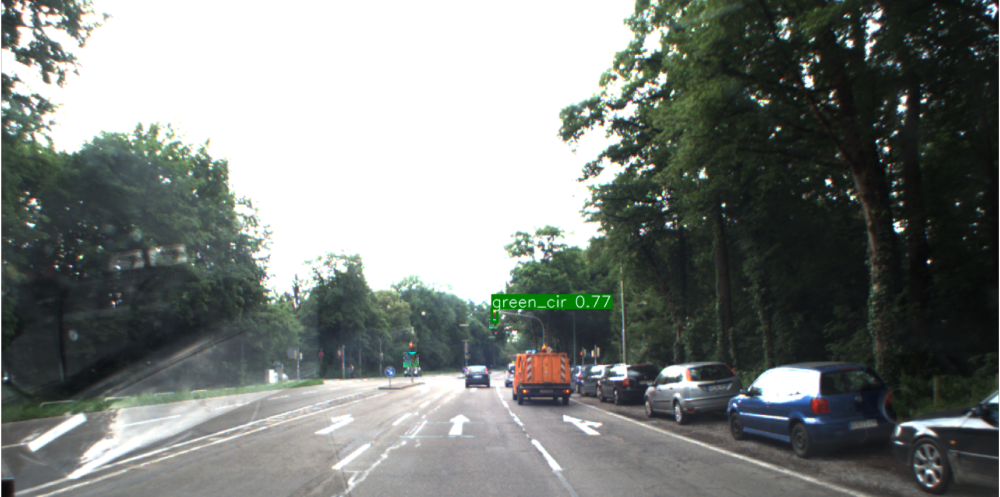}
  	\caption*{\texttt{HDTLR}}
\end{subfigure}

\begin{subfigure}[t]{0.72\linewidth}
    \includegraphics[width=\textwidth]{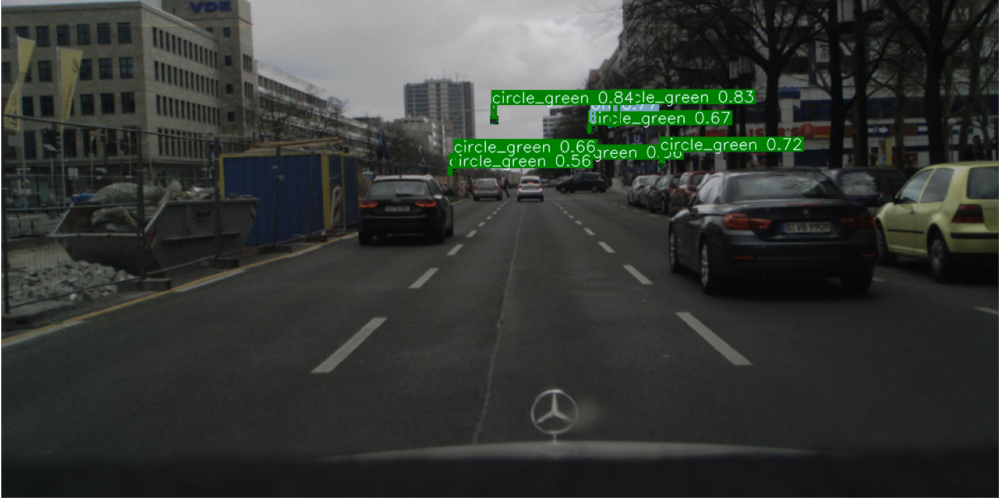}
  	\caption*{\texttt{DTLD}}
\end{subfigure}
    \caption{Universal attacks, predictions by YOLOv7x.}
    \label{fig:attack-datasets}
\end{figure}

\textbf{Impact of loss components:} we used $\alpha = 1$ and $\gamma = 0.8$. $\mathcal{L}_{\text{cls}}$ and $\mathcal{L}_{\text{tv}}$ were sufficient for generating patches with smooth pixel transitions (see Figure~\ref{fig:patches}) leading to stable label flips without object fabrication (see Figure~\ref{fig:attack-datasets}).

\textbf{Vulnerability across the datasets}: images from the \texttt{BSTLD} dataset were the easiest to attack, which a smaller number of classes can explain. For \texttt{HDTLR} and \texttt{LISA}, object vanishing and detection of non-existent objects on the patch were observed. If several traffic lights are present in an image, an attack on a larger object was more successful due to patch rescaling effects.

\textbf{Vulnerability across models}: YOLOv7 models have demonstrated a slightly larger vulnerability. YOLOv8 benefits from improved training techniques, such as better augmentation and optimization strategies, and includes stronger regularization, making it more resilient to adversarial attacks.


\begin{figure}[h]
\centering
    \includegraphics[width=0.72\linewidth]{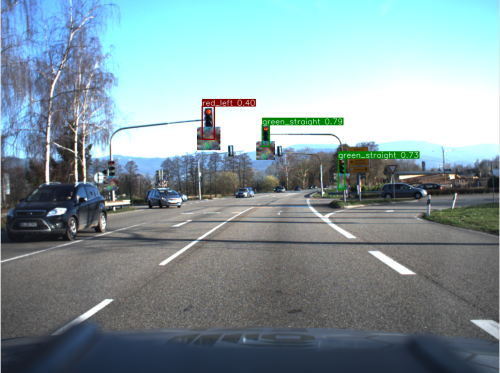}
    \caption{Example of a pictogram flipping attack for arrows. Predictions by YOLOv7 on \texttt{HDTLR} dataset.}
    \label{fig:arrows}
\end{figure}

\textbf{Attacks on arrow labels:} because various arrow pictograms exhibit high similarity, the \textit{red-arrow}$\to$\textit{green-arrow} patch attack leads to misclassification not only of red arrows but also of other green arrow types (see Figure~\ref{fig:arrows}).

\clearpage
\newpage
\section{Real-world Evaluation with a Mobile Traffic Light}
For tests in the lab, we used a mobile traffic light, typically used to regulate one-way alternating traffic at construction sites. It only has red, yellow, and green states without arrows.


\subsection{Experimental Setup}
We took images of the traffic light from a distance of 2-10m. The traffic light in images is larger than in the datasets used in Section~\ref{ref:sec_digital}; therefore, it could only be reliably detected from a distance of more than 6 m.

To select a model and train a patch, we labeled 388 images of this traffic light at different outdoor scenes, with an even number of pictures with red and green traffic light states. We evaluated several models on this dataset. Models trained on the \texttt{BSTLD} dataset have demonstrated the best performance on clean data compared to other datasets. Out of \texttt{BSTLD} models, YOLOv7 was selected for further experiments (see Table~\ref{tab:mobile-performance}). We used $\alpha = 1$, $\beta = 2$, $\gamma = 5$, and $\delta = 0.0002$ for patch generation.

\begin{table}[h]
\caption{Evaluation of BSTLD models on the test subset of mobile traffic light data, \textbf{no attack}.}
\label{tab:mobile-performance}
\begin{center}
    \begin{tabular}{|r|c| c|c| }
    \hline
    \textbf{Model} & \textbf{mAP50} &  \textbf{$AP_{red}$} &  \textbf{$AP_{green}$} \\ \hline
    YOLOv7 & \textbf{0.996} & \textbf{0.997} & \textbf{0.996} \\
    YOLOv7x & 0.794 & 0.591 & \textbf{0.996} \\
    YOLOv8m & 0.935 & 0.876 & 0.995 \\
    YOLOv8x & 0.853 & 0.711 & 0.995 \\
    RT-DETR-L & 0.993 & 0.992 & 0.995 \\
    RT-DETR-X & 0.603 & 0.232 & 0.973 \\
    \hline
    \end{tabular}
\end{center}
\end{table}

\subsection{Analysis}
We evaluate attack success visually by placing generated patches under the traffic light and taking images at different outdoor locations.

\textbf{Patch size}: three patch sizes were evaluated: 1.5, 2, and 2.5 times the width of the light. The small patch (45$\times$45cm) could not successfully suppress red detections as distance increased. The medium (60$\times$60cm) and large (75$\times$75cm) patches performed similarly regardless of distance.

\begin{figure}[t]
    \centering
    \begin{subfigure}[t]{\linewidth}
        \includegraphics[width=0.45\textwidth]{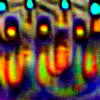}
        \includegraphics[width=0.45\textwidth]{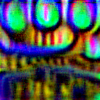}
      	\caption{Patches used in the attacks.}
    \end{subfigure}
    \begin{subfigure}[t]{\linewidth}
        \includegraphics[width=0.45\textwidth]{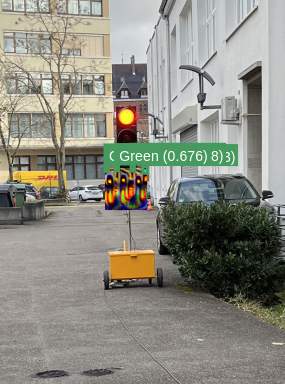}
        \includegraphics[width=0.45\textwidth]{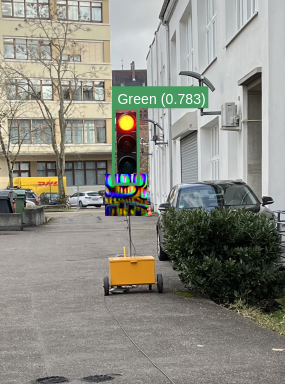}
      	\caption{Evaluation in the digital setting.}
    \end{subfigure}
    \begin{subfigure}[t]{\linewidth}
        \includegraphics[width=0.45\textwidth]{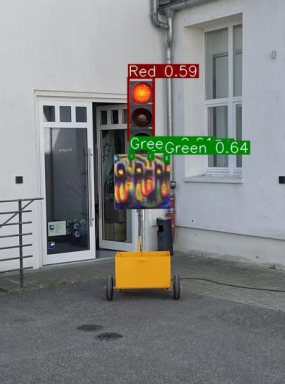}
        \includegraphics[width=0.45\textwidth]{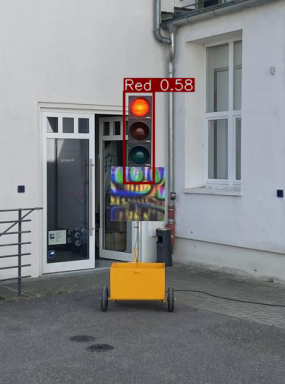}
      	\caption{Evaluation with a printed patch.}
    \end{subfigure}
    \begin{subfigure}[t]{\linewidth}
        \includegraphics[width=0.45\textwidth]{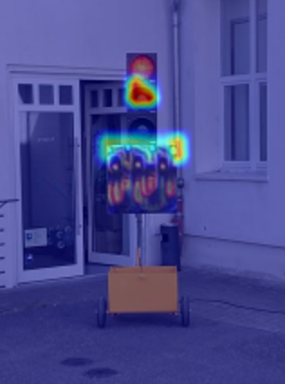}
        \includegraphics[width=0.45\textwidth]{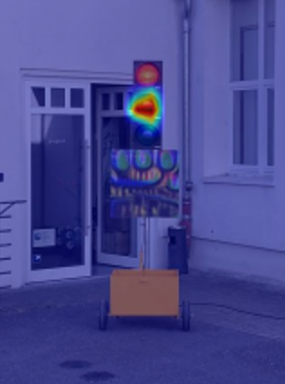}
      	\caption{GradCAM analysis of real-world attacks.}
    \end{subfigure}    
    \caption{Experiments with a mobile traffic light system. The left column uses a patch trained without, and the right column - with $\mathcal{L}_{\text{bbox}}$.}
    \label{fig:tl-lab}
\end{figure}

\textbf{Impact of $\mathcal{L}_{\text{bbox}}$}:  unlike in experiments with the datasets, BBoxes are considerably large. We used $\mathcal{L}_{\text{bbox}}$ to ensure correct BBox prediction (see Figure~\ref{fig:tl-lab}). While using $\mathcal{L}_{\text{bbox}}$ led to a successful attack in the digital setting, it resulted in suppressing red detections without the fabrication of green lights in the real-world evaluation. As GradCAM~\cite{selvaraju2017gradcam}\footnote{https://github.com/jacobgil/pytorch-grad-cam} analysis shows, a patch without BBox constraints had an area of high green saliency at the top, whereas a patch with $\mathcal{L}_{\text{bbox}}$ has no salient regions, indicating that BBox constraints prevent fabrication of detections but do not affect the suppression of red detections.

\textbf{Impact of $\mathcal{L}_{\text{green\_sup}}$}: the usage of a green suppression loss led to the generation of smaller green blobs and subsequently less object fabrication effects.

\clearpage
\newpage
\section{Real-world Evaluation in the Test Area}
Finally, the proposed attacks were evaluated using an automated vehicle with a running traffic light perception pipeline in a test area for autonomous driving. 

\subsection{Experimental Setup}
To assess patch performance in real-world traffic scenes, we evaluated attacks at Campus East, a restricted Karlsruhe Institute of Technology (KIT) area. As a part of the Test Area Autonomous Driving Baden-Württemberg~\footnote{\url{https://taf-bw.de/}}, it is dedicated to early rapid prototyping and vehicle tests with about 2~km of road and 5000~m$^2$ free space. The evaluation was performed with our research vehicle \textit{CoCar NextGen}~\cite{heinrich2024cocarnextgen} based on an Audi Q5. We used our traffic light perception framework ATLAS~\cite{polley2025atlas}, which integrates YOLO-based object detections with temporal, spatial, and map-derived information to determine final traffic light states relevant to the autonomous vehicle's decision-making. 

The ATLAS framework uses two parallel camera streams: the first for the front-medium camera, whereas the second alternates between the front-wide and front-tele cameras. A separate YOLO model processes each camera stream. 


Because of good performance during the road tests in our previous work~\cite{polley2023tld}, we used the YOLOv8x model trained on DTLD data. To train patches, we used images from the ATLAS dataset ~\cite{polley2025atlas}\footnote{\url{https://url.fzi.de/ATLAS}}, showing three different traffic lights at Campus East. Enforcing the correct BBox coordinates with  $\mathcal{L}_{\text{bbox}}$ and reducing the number of green pixels with $\mathcal{L}_{\text{green\_sup}}$ have not resulted in successful attacks. Furthermore, minor discrepancies between the true and the detected BBox coordinates are allowed within the ATLAS pipeline as long as they stay within a certain tolerance to the traffic light coordinates stored in the HD map.

The evaluation was performed while driving autonomously, whereas the distance to a traffic light varied between 1 and 7 m.   

\subsection{Analysis}
\textbf{Attack success:} differently from our previous experiments on datasets and with a mobile traffic light, where the decision-making process of a single YOLO model was compromised, attacking the ATLAS pipeline is more challenging due to its complexity. To alter the output driving decision of the pipeline, the attack should consistently manipulate multiple frames, preserving the altered traffic light label's consistency and directionality (e.g., arrows). The ATLAS pipeline inherently filters out object fabrication attacks. Also, object vanishing attacks always default to an \textit{unknown} state. Furthermore, for traffic intersections with multiple traffic lights relevant for the ego-vehicle, attacking only one traffic light proved ineffective since the unaffected detection overrides the compromised one, leading ATLAS to select the correct driving maneuver. At single-light intersections, attack effectiveness depended significantly on environmental conditions. Attacks only marginally reduced confidence scores under bright, clear sky conditions without achieving label flips. Conversely, in overcast conditions, label-flipping became possible but remained inconsistent. 

\begin{figure}[t]
\centering
    \includegraphics[width=0.45\linewidth]{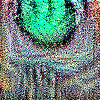}
    \includegraphics[width=0.45\linewidth]{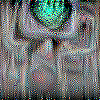}
    \caption{Examples of patches generated for real-world experiments with a stationary traffic light. Left: patch generated without  $\mathcal{L}_{\text{tv}}$. Right: patch generated with $\gamma$=1. }
    \label{fig:real-patches}
\end{figure}

\begin{figure}[t]
\centering
    \begin{subfigure}[t]{\linewidth}
     \includegraphics[width=0.495\columnwidth]{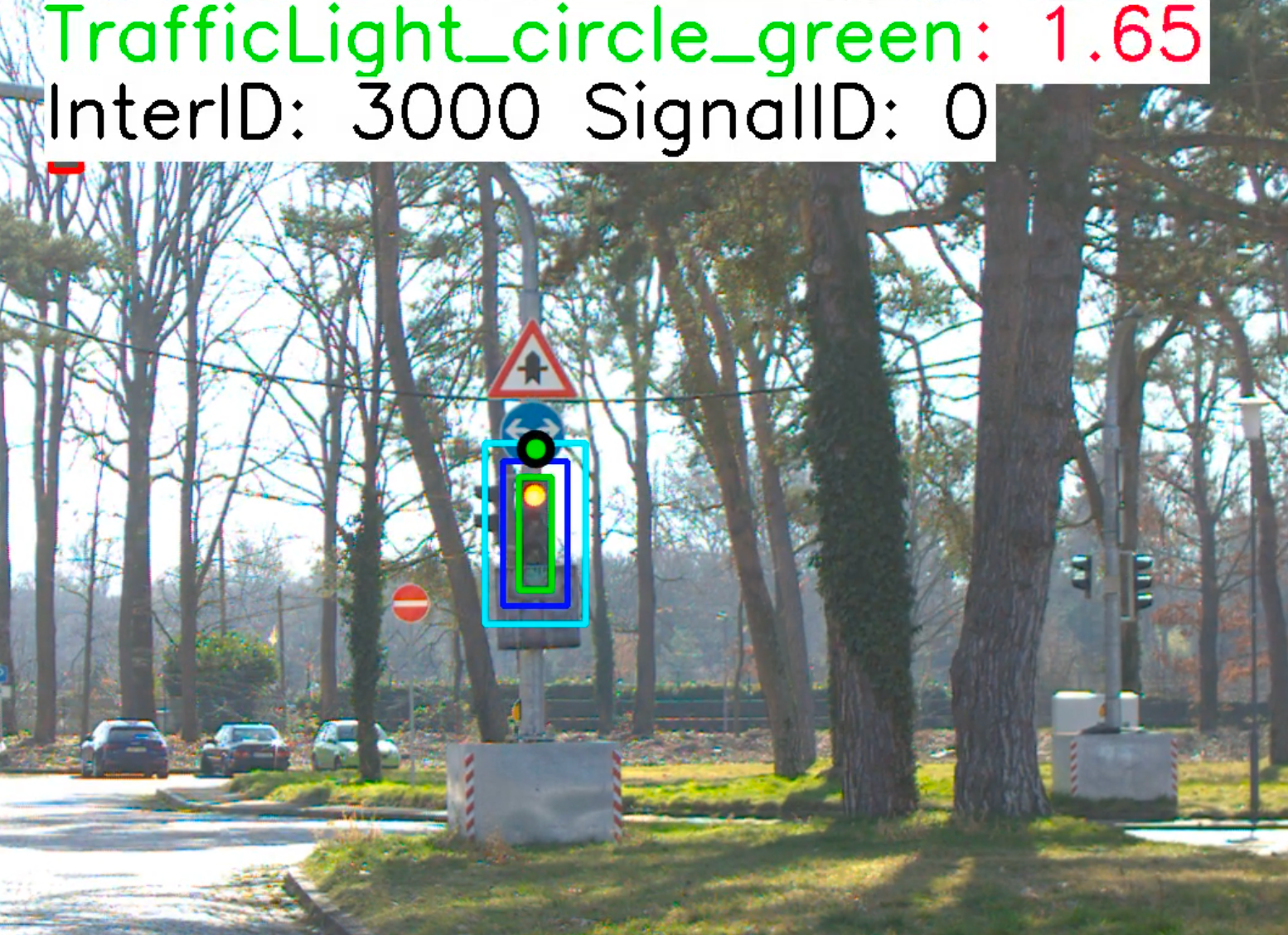}
        \includegraphics[width=0.495\columnwidth]{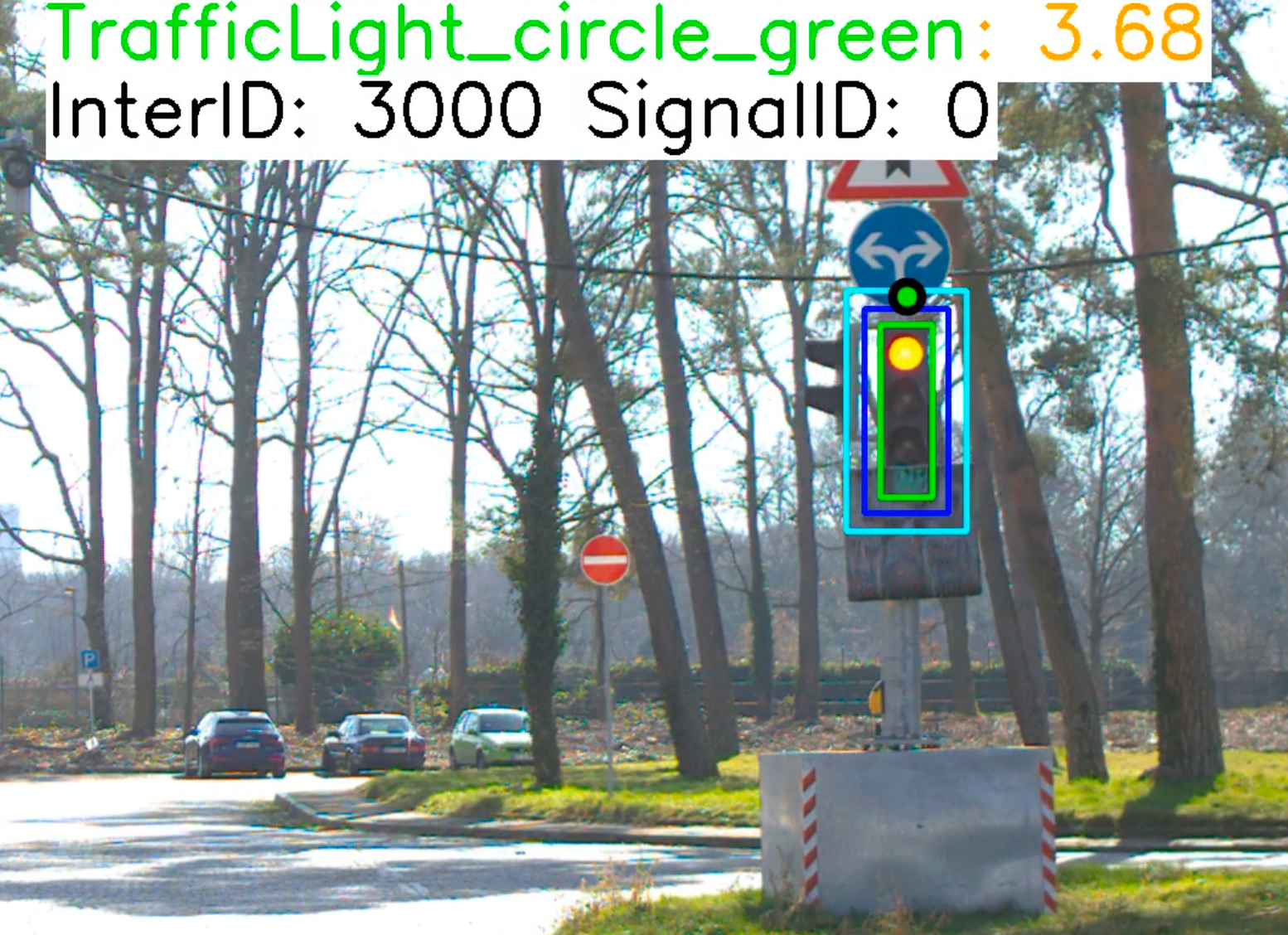}
    \caption{The confidence of the final driving decision is lower for smaller distances, color-coded in red and orange.}
    \end{subfigure}
    \hspace{5mm}
    \begin{subfigure}[t]{\linewidth}
        \includegraphics[width=0.495\columnwidth]{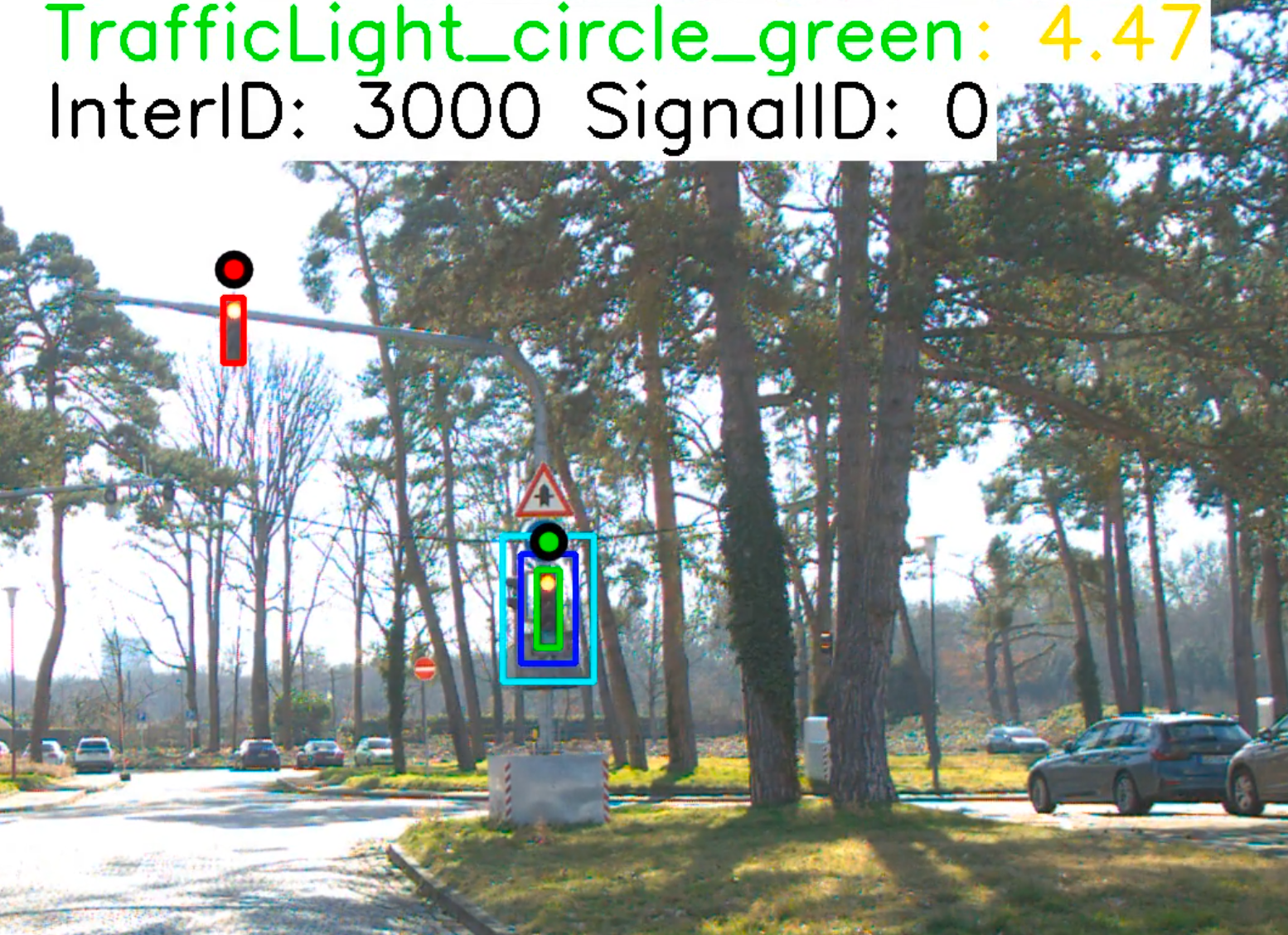}
        \includegraphics[width=0.495\columnwidth]{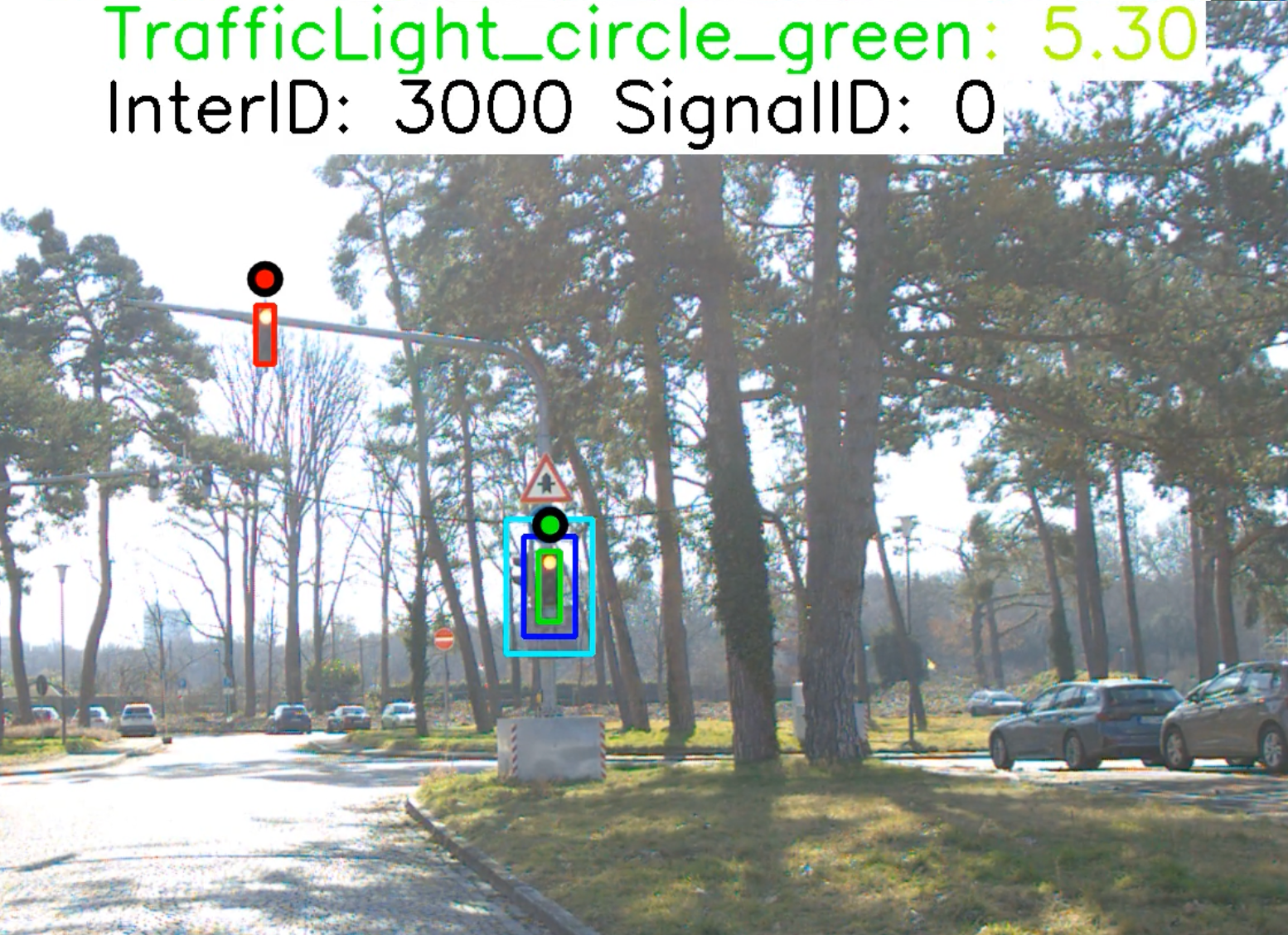}
    \caption{For larger distances to the camera, the patch leads to a larger confidence of the driving decision, color-coded in yellow and green.}
        
    \end{subfigure}
    \caption{Experiments with a stationary traffic light, front-medium camera images of a red traffic light. Text on top is the ATLAS output: the detected class (here: \textit{TrafficLight\_circle\_green}) and the color-coded confidence of the final driving decision. Blue and teal BBoxes denote that ATLAS identifies the traffic light as relevant to the ego vehicle and incorporates its classification into the driving decision-making process. The green BBox shows the final driving decision of the ATLAS pipeline.}
    \label{fig:campus-ost-attacks}
\end{figure}

\textbf{Distance to the camera:} successful \textit{red}$\to$\textit{green} label flips have been reliably achieved at medium distances, causing the vehicle to continue with its current speed and no deceleration, thus violating the red light (see Figure~\ref{fig:campus-ost-attacks}). However, at distances below 7 m, the ATLAS module occasionally recovered the correct state, causing rapid oscillations between \textit{red} and \textit{green} detections. This led to a cycle of maintaining speed and abrupt full braking, creating highly unsafe driving dynamics.  At distances below 2 m, the true \textit{red} traffic light state was consistently recovered. However, delayed recognition caused the vehicle to halt dangerously past the stop line in the intersection due to inertia.

\textbf{Patch size:} out of the evaluated sizes (40$\times$40cm, 60$\times$60cm, and 80$\times$80cm), the latter led to the best performance. Larger patches are unfeasible for a realistic attack.

\textbf{Patch appearance:} The resulting patches (see Figure~\ref{fig:real-patches}) exhibit a stronger resemblance to traffic lights than those generated solely on the datasets (see Figure~\ref{fig:patches}). 

In summary, attacking an autonomous vehicle's traffic light perception pipeline is significantly more challenging than attacking a single CNN. While successful attacks were observed during fully autonomous driving, especially at larger distances from the patch, the attacks were either weaker or not functioning for smaller distances.

\section{Conclusion}
In this work, we have evaluated adversarial patch-based attacks on camera-based traffic light detection models. We proposed a threat model to attack traffic light detection with an adversarial patch placed directly under a traffic light and resized correspondingly. We trained a patch using PGD while updating a patch for each traffic light in an image. We additionally used total variation loss to ensure smooth transitions between neighboring pixels and proposed a further loss component to penalize large green areas on a patch. Patches generated with the proposed approach have demonstrated high attack strength over all four datasets in the digital settings. The evaluation in the real world has shown that a patch can also trigger object vanishing and object fabrication effects, which can be reduced with the proposed losses. Apart from attacking single detection models, we have also evaluated the impact of the proposed attack on a modularized traffic light detection pipeline, relying on HD map data and including real-time association and decision-making. Our experiments with a research vehicle driving fully autonomously have shown that the proposed attack leads to successful \textit{red}$\to$\textit{green} label flips in this setting. 

The proposed attack can further be enhanced in the future to ensure more stable label flipping. Furthermore, possibilities to make the patch look inconspicuous might be explored.

\section*{Acknowledgment}

This work was supported by funding from the Topic Engineering Secure Systems of the Helmholtz Association (HGF) and by KASTEL Security Research Labs (46.23.03).

{\small
\bibliographystyle{IEEEtran}
\bibliography{references}
}

\end{document}